\documentclass[letterpaper]{article}
\usepackage[preprint]{aaai2027}
\usepackage[hyphens]{url}
\usepackage{graphicx}
\usepackage{natbib}
\usepackage{caption}
\usepackage{amsmath,amssymb,amsfonts}
\usepackage{algorithm}
\usepackage{algorithmic}
\usepackage{booktabs}
\usepackage{pifont}
\usepackage{xspace}
\usepackage{multirow}


\newcommand{\cmark}{\ding{51}}
\newcommand{\xmark}{\ding{55}}

\newcommand{\answerTODO}[1][]{\textcolor{red}{\bf [TODO]}}
\newcommand{\rev}[1]{#1}

\title{FedSAP: Federated Learning with Structured Adaptive Partitioning for Multi-Domain Heterogeneous Edge Devices}

\author{
Wentao Yue, Tianyou Lai, Hongji Li,\\
Qingyu Mao, Qilei Li
}
\affiliations{}

\begin{document}

\maketitle

\begin{abstract}
\rev{Federated learning (FL) on heterogeneous edge devices must jointly accommodate unequal resource budgets and domain-shifted local data. Existing resource-adaptive methods decide how much of a model each client trains but not where retained capacity should reside or how it should be shared, whereas federated domain-generalization methods usually assume a shared full architecture. Uniform compression can therefore discard high-utility channels, and a single aggregation path can mix transferable features with domain-sensitive updates. We propose \textbf{FedSAP}, a domain-aware heterogeneous FL framework that casts structured pruning as budget-constrained tri-state channel allocation. FedSAP converts each keep ratio into non-uniform layer budgets, assigns stable channels to a \emph{Global} pool, useful domain-sensitive channels to pseudo-domain-specific \emph{Private} pools, and low-utility channels to a \emph{Dropped} state. This partition lets broadly useful features benefit from cross-client pooling while isolating domain-sensitive updates from incompatible clients. Domain-Guided Assignment infers pseudo-domains from shallow-gradient similarity, while Type-Matched Aggregation restricts each channel to its intended sharing scope. Across three random seeds, FedSAP reaches $76.00\%$ and $72.67\%$ mean global accuracy on Digits and Office-Caltech, exceeding the strongest baseline by $1.70$ and $4.92$ percentage points while supporting client pruning ratios of up to $80\%$ across heterogeneous clients.}

\end{abstract}

\section{Introduction}
\label{sec:intro}

\begin{figure*}[t]
    \centering
    \begin{minipage}[t]{0.48\textwidth}
        \centering
        \includegraphics[width=0.8\linewidth]{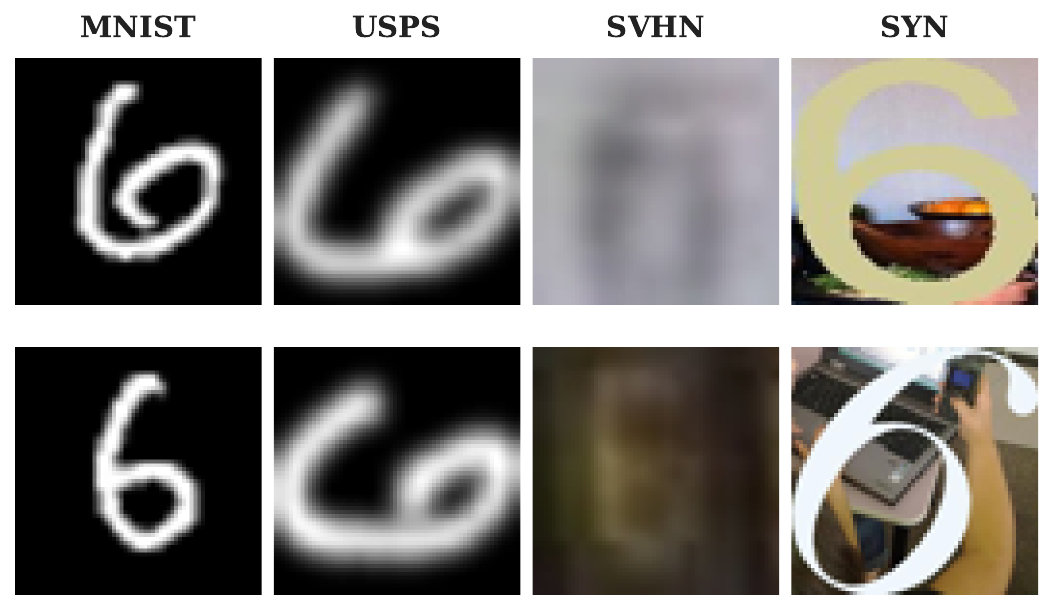}
        \vspace{0.2em}
        \centerline{\small (a) Digits}
    \end{minipage}
    \hfill
    \begin{minipage}[t]{0.48\textwidth}
        \centering
        \includegraphics[width=0.8\linewidth]{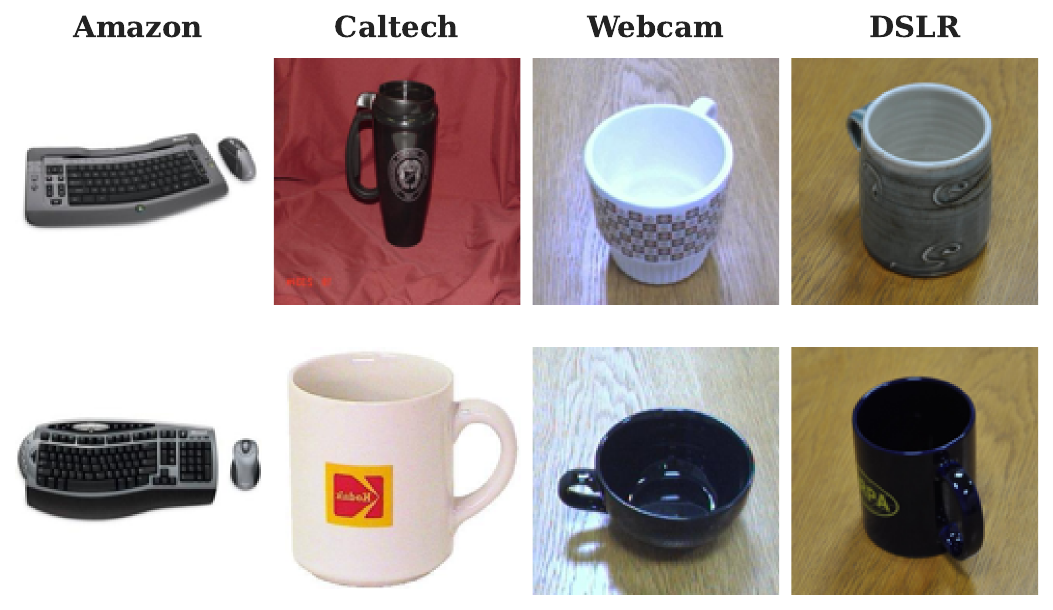}
        \vspace{0.2em}
        \centerline{\small (b) Office-Caltech}
    \end{minipage}
    \caption{
    Representative domain shifts.
Digits spans handwritten, street-view, and synthetic domains; Office-Caltech spans different backgrounds, viewpoints, resolutions, and camera conditions.
    }
    \vspace{-1.0em}
    \label{fig:domain_shift_examples}
\end{figure*}

\rev{Federated learning (FL) trains a shared model across edge devices without centralizing their raw data~\citep{mcmahan2017communication,huang2024federated}. Real deployments, however, couple two forms of heterogeneity~\citep{yue2026review,ye2023heterogeneous}. Devices differ in computation, memory, and energy budgets~\citep{ilhan2023scalefl}, so a model that fits a server-grade client may be infeasible on a low-end device~\citep{yang2026tackling}. Their data also arise from different environments~\citep{li2025federated}: digit images vary from grayscale handwriting to street-view crops and synthetic renderings, while object images vary in camera, background, resolution, and style (Figure~\ref{fig:domain_shift_examples}). Consequently, local updates encode both task knowledge and domain bias; indiscriminate aggregation can transfer the former while amplifying the latter.}

\rev{The two problems cannot be solved independently. Resource-adaptive FL assigns smaller subnetworks to weak devices~\citep{bouacida2021adaptivefederateddropout,diao2021heterofl,jiang2022fedmp}, but efficiency-oriented pruning does not identify which retained features should transfer across domains. Federated domain-generalization methods improve robustness to distribution shifts~\citep{caldarola2022fedsr,huang2023fpl,pourpanah2025fedgala}, but usually require every client to train the same full model. DapperFL~\citep{jia2024dapperfl} connects these settings through model-fusion pruning and domain-adaptive regularization, enabling differently compressed clients under domain shift. Yet a compression ratio specifies only \emph{how much} capacity remains; it does not specify \emph{where} that capacity should reside or \emph{with whom} each retained feature should be shared.}

\rev{This missing allocation decision creates three concrete failure modes. A uniform layer policy can spend scarce capacity on low-utility layers; a single aggregation path can overwrite domain-sensitive channels with incompatible updates; and shape reconstruction can align tensors without determining which clients should share those channels. Addressing coupled heterogeneity therefore requires a joint decision over layer budget, channel role, and sharing scope.}

\rev{We therefore propose \textbf{FedSAP} (\underline{\textbf{Fed}}erated learning with \underline{\textbf{S}}tructured \underline{\textbf{A}}daptive \underline{\textbf{P}}artitioning), a domain-aware pruning framework that turns a fixed client budget into a structured allocation across layers and sharing roles. FedSAP first assigns more active channels to layers with higher estimated utility. It then partitions channels into three states: stable and useful channels are \emph{Global}, useful but domain-sensitive channels are \emph{Private}, and low-utility channels are \emph{Dropped}. Domain-Guided Assignment (DGA) infers pseudo-domain groups from shallow-gradient similarity, and Type-Matched Aggregation shares Global channels across all clients while restricting Private-channel aggregation to the corresponding pseudo-domain group. Under client pruning ratios from $0\%$ to $80\%$, FedSAP improves mean global accuracy on both evaluated benchmarks. The gain is an aggregate trade-off rather than uniform dominance: some individual domains decline, a point we quantify in the Main Results subsection and analyze further in the supplementary material.}
\newpage
\begin{itemize}
    \item \rev{We formulate resource adaptation under domain shift as a joint allocation problem: a client-level budget must determine both where capacity is retained and how each retained channel is shared.}

    \item \rev{FedSAP combines adaptive layer-wise budgets with tri-state channel partitioning, assigning channels to \emph{Global}, \emph{Private}, or \emph{Dropped} states according to utility, cross-domain stability, and client constraints.}

    \item \rev{DGA and Type-Matched Aggregation align each active channel with an estimated sharing scope. Experiments over three seeds show higher global mean accuracy under heterogeneous compression and report the domain-level trade-offs. The supplement provides budget--aggregation and domain-wise budget experiments, full sensitivity results, conditional theory, and privacy analysis.}
\end{itemize}

\section{Related Work}
\label{sec:related}

\noindent\textbf{Resource-adaptive federated learning (FL).}
FL on edge devices is challenged by large heterogeneity in computational, memory, and communication resources across clients. A common solution is to assign clients subnetworks of different sizes. HeteroFL~\citep{diao2021heterofl} trains width-scaled local models under a shared global architecture, while FjORD~\citep{horvath2021fjord} uses ordered dropout to obtain nested subnetworks for clients with different budgets. FedRolex~\citep{alam2022fedrolex} rotates partial subnetworks across rounds to improve the coverage of global parameters. Others construct personalized or importance-aware submodels, such as FedP3~\citep{yi2024fedp3}, FIARSE~\citep{wu2024fiarse}, and pFedGate~\citep{chen2023pfedgate}. These methods improve the practicality of FL in the presence of system heterogeneity, but their primary focus is efficiency or personalization in conventional non-IID settings. They generally overlook explicitly model domain shifts in which clients differ in visual style, background, viewpoint, or acquisition conditions. Thus, a compact model optimized for one domain may discard channels that are important for another domain.

\noindent\textbf{FL under domain shift.}
Domain generalization aims to learn representations transferable to unseen domains. In FL, this problem is even harder because source domains are distributed across clients, and raw data cannot be centralized. Existing federated domain generalization methods mainly use representation regularization, prototype alignment, contrastive learning, or gradient-based aggregation. FedSR~\citep{caldarola2022fedsr} regularizes latent representations to reduce domain-sensitive variation, and FPL~\citep{huang2023fpl} aligns class-level semantics through prototypes. FedDAR~\citep{zhong2023feddar} learns a shared encoder with domain-wise prediction heads, while recent methods such as FedOMG~\citep{nguyen2025federated} and FedGaLA~\citep{pourpanah2025fedgala} exploit gradient signals to improve domain robustness without server-side data. Although these methods address domain shifts, they are developed under an architecture-homogeneous full-model setting, assuming that all clients train the same backbone with identical capacity. This limits their applicability to model-size heterogeneous edge systems, where weak clients may only support compact subnetworks. Our work instead studies domain robustness under client-specific pruning budgets.

\noindent\textbf{Federated pruning with domain-aware aggregation.}
Model pruning is widely used in FL to reduce computation and communication costs. FedDrop~\citep{bouacida2021adaptivefederateddropout}, PruneFL~\citep{jiang2022prunefl}, and FedMP~\citep{jiang2022fedmp} generate compact models under client resource constraints, but mainly treat pruning as a compression problem. They do not explicitly decide whether the retained capacity should support globally transferable or domain-specific knowledge. DapperFL~\citep{jia2024dapperfl} is the closest work to our setting. It combines Model Fusion Pruning with Domain Adaptive Regularization and reconstructs heterogeneous pruned models before aggregation, so clients with different pruning ratios can participate under domain shifts. However, its pruning ratio mainly controls \emph{how much capacity each client retains, rather than where this limited capacity should be placed across layers.} As a result, layers with different utility may still be compressed under a similar budget pattern. In addition, all retained channels follow the same aggregation path after reconstruction, although some channels may encode globally transferable features while others capture domain-sensitive patterns. This makes it difficult to preserve both local specialization and global transfer simultaneously. CA-HFP~\citep{hu2026cahfp} uses curvature-aware scores and model reconstruction to stabilize heterogeneous pruning, but its focus is on pruning robustness under statistical heterogeneity. It does not explicitly model pseudo-domains or assign retained channels to domain-aware aggregation paths. \rev{In contrast, FedSAP allocates active channels non-uniformly across layers, assigns them to \emph{Global} or \emph{Private} aggregation paths, and \emph{Drops} low-utility channels to satisfy each client's capacity constraint during local training.}


\section{Methodology}

\subsection{Problem Formulation and Overview}

\begin{figure*}[t]
    \centering
    \includegraphics[width=0.8\textwidth]{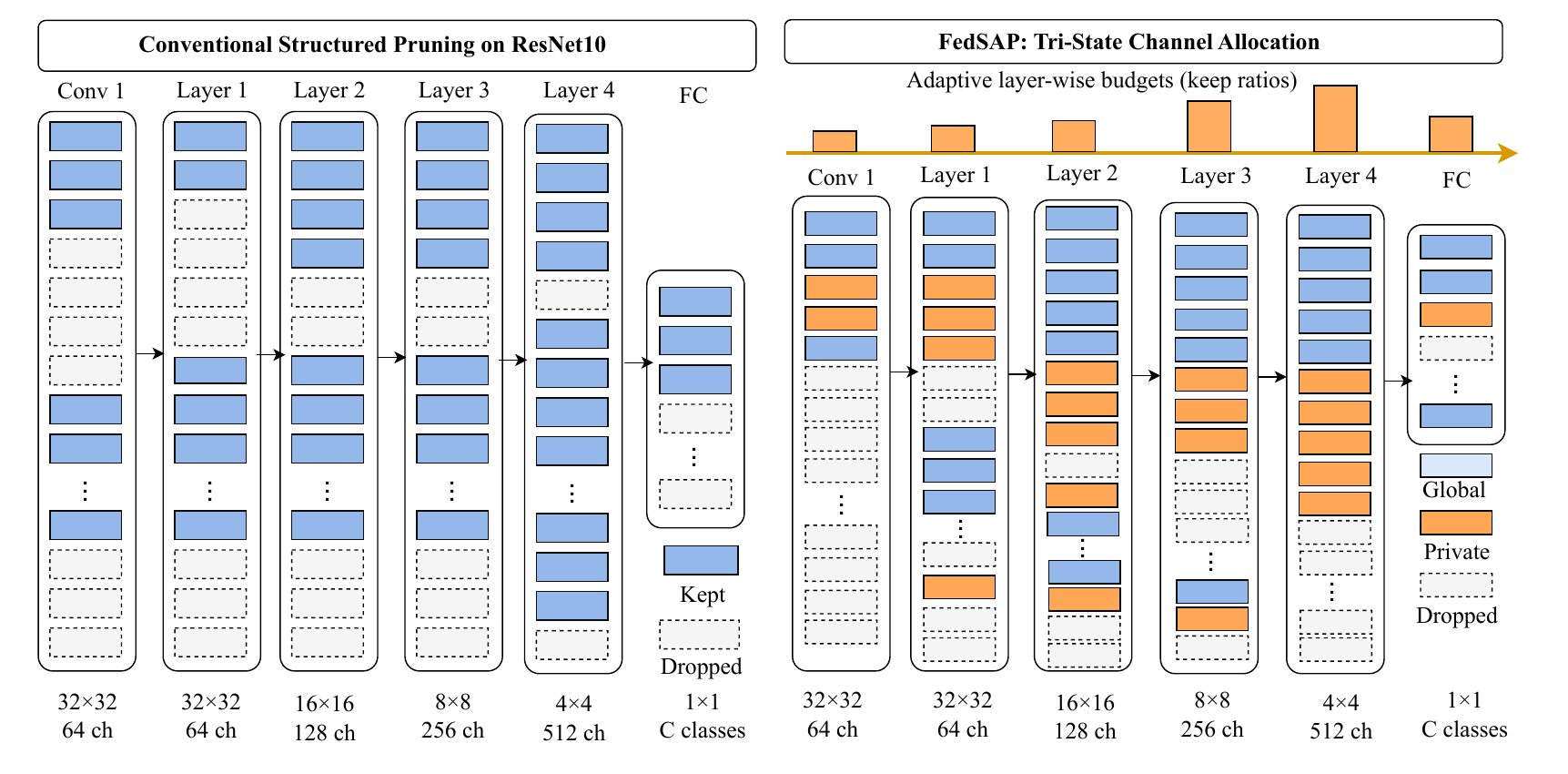}
    \caption{
Conventional pruning versus FedSAP on ResNet10.
Conventional pruning keeps one channel set, whereas FedSAP distributes the client budget across layers and assigns channels to \emph{Global} or \emph{Private} type-matched paths.
}
    \label{fig:compare_pruning}
\end{figure*}

We consider a federated learning system with $N$ clients. Each client $i$ owns a private dataset $\mathcal{D}_i$ sampled from an unknown local distribution. The global objective is
\begin{equation}
\label{eq:fl_objective}
    \min_{\mathbf{W}}
    \sum_{i=1}^{N} p_i \mathcal{L}_i(\mathbf{W}; \mathcal{D}_i),
    \quad
    p_i=\frac{|\mathcal{D}_i|}{\sum_{j=1}^{N}|\mathcal{D}_j|},
\end{equation}
where $\mathbf{W}$ denotes the global model and $\mathcal{L}_i$ is the empirical loss on client $i$. We study a practical edge setting where clients differ in both system capacity and data domain. Following the heterogeneous device protocol in prior work~\citep{jia2024dapperfl}, each client is assigned a pruning ratio $\rho_i \in \{0,0.2,0.4,0.6,0.8\}$ according to its resource budget. \rev{The corresponding target keep ratio is $\kappa_i=1-\rho_i$. Figure~\ref{fig:compare_pruning} contrasts conventional structured pruning with FedSAP under this client-level budget.} Conventional pruning keeps one retained channel set, whereas FedSAP separates layer-wise capacity allocation from channel aggregation roles, allowing globally transferable and domain-sensitive channels to follow different paths.

\begin{figure*}[t]
    \centering
    \includegraphics[width=0.8\textwidth]{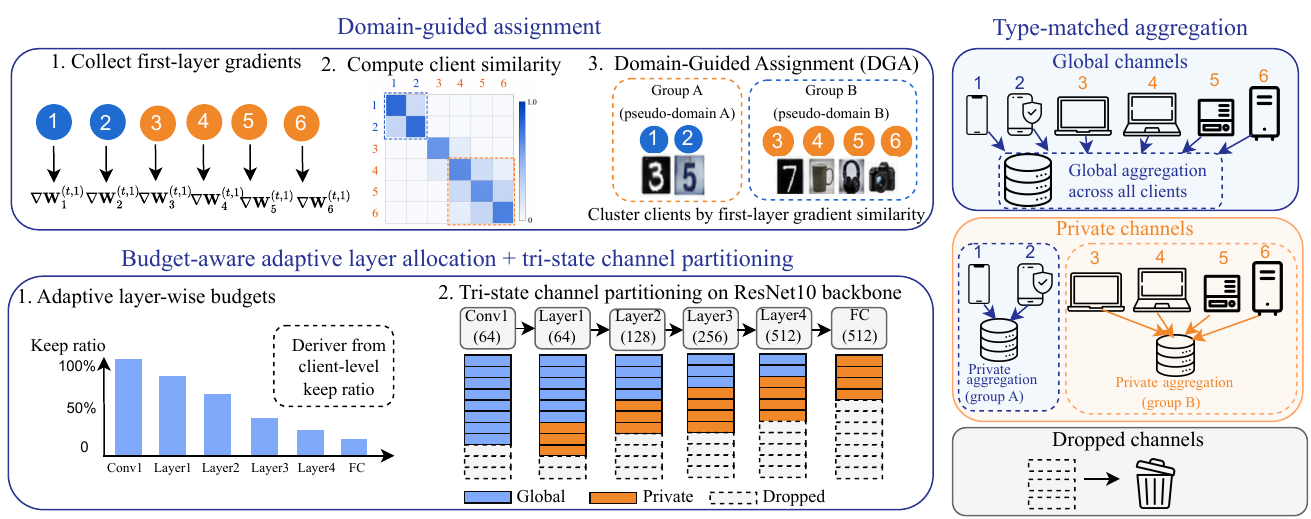}
    \vspace{-0.5em}
   \caption{
Overview of FedSAP: the server discovers pseudo-domain groups, allocates client-level budgets across layers, assigns channels to \emph{Global}, \emph{Private}, or \emph{Dropped} states, and aggregates channels through matched sharing scopes.
}
    \label{fig:fedsap_flow}
    \vspace{-0.8em}
\end{figure*}

As illustrated in Figure~\ref{fig:fedsap_flow}, FedSAP treats $\kappa_i$ as a client-level capacity budget rather than a fixed pruning rule for every layer. For a model with $L$ convolutional layers, let $C_\ell$ be the number of output channels in layer $\ell$. FedSAP first converts the client-level budget into adaptive layer-wise keep ratios $\{\kappa_{i,\ell}^{(t)}\}_{\ell=1}^{L}$, which satisfy the global budget constraint
\begin{equation}
\label{eq:budget_constraint}
    \frac{
    \sum_{\ell=1}^{L} C_\ell \kappa_{i,\ell}^{(t)}
    }{
    \sum_{\ell=1}^{L} C_\ell
    }
    =
    \kappa_i .
\end{equation}
Thus, different layers can receive different active-channel budgets under the same client-level compression ratio. Layers with higher estimated utility can retain more channels, whereas less useful layers can undergo more pruning.

\rev{Within each layer, FedSAP assigns every channel to a Global, Private, or Dropped state according to its utility and intended sharing scope during training:}
\begin{equation}
\label{eq:tri_state_mask}
    \mathcal{M}_{i,\ell}^{(t)}(c) \in \{\mathrm{G}, \mathrm{P}, \mathrm{D}\},
    \quad c \in \{1,\ldots,C_\ell\}.
\end{equation}
Here, $\mathrm{G}$ denotes a Global channel shared across all clients, $\mathrm{P}$ denotes a Private channel shared only within a pseudo-domain group, and $\mathrm{D}$ denotes a Dropped channel that remains inactive under the client budget. The active channels, i.e., those assigned to $\mathrm{G}$ or $\mathrm{P}$, must respect the layer-wise budget $\kappa_{i,\ell}^{(t)}$. This formulation poses two questions simultaneously: \emph{1) where should the limited channel capacity be allocated across layers; and 2) how should each active channel be shared among clients?}

FedSAP answers these questions through three coupled operations. Domain-Guided Assignment estimates pseudo-domain groups from first-layer gradient similarity without domain labels. Selective Global Protection assigns stable important channels to the Global pool and important domain-sensitive channels to Private pools. Type-matched aggregation then averages Global channels across all clients and Private channels only within the corresponding pseudo-domain group. \rev{This turns a fixed client pruning ratio into layer-adaptive, role-aware channel allocation under heterogeneous domain shifts and resource constraints.}

\label{sec:method}

\subsection{Domain-Guided Assignment}


Motivated by recent studies showing that gradient information reflects domain shift and can guide federated domain adaptation or generalization~\citep{jiang2023principled,pourpanah2025fedgala}, FedSAP uses layer gradient similarity as a lightweight signal for pseudo-domain discovery.
At Domain-Guided Assignment (DGA) update rounds, after local training, each client reports the gradient of the first convolutional layer, denoted by $\mathbf{g}_i^{(t)}$, to the server. The server computes the pairwise cosine similarity
\begin{equation}
\label{eq:dga_similarity}
    S_{ij}^{(t)}
    =
    \frac{
    \langle \mathbf{g}_i^{(t)}, \mathbf{g}_j^{(t)} \rangle
    }{
    \|\mathbf{g}_i^{(t)}\|_2 \|\mathbf{g}_j^{(t)}\|_2
    } .
\end{equation}
The similarity matrix $\mathbf{S}^{(t)}$ defines pseudo-domain groups
$\{\mathcal{G}_1^{(t)},\ldots,\mathcal{G}_K^{(t)}\}$. 
\rev{The pseudo-domain grouping is updated every $R_d$ communication rounds to limit communication and reduce assignment changes caused by noisy early gradients. We use the first-layer gradient as a deliberately lightweight proxy because it responds strongly to texture, background, color, and stroke statistics. This choice is conditional rather than universal: it is most appropriate when domain identity is expressed through low-level appearance and can fail when domains are defined mainly by high-level semantics. The Limitations section and the supplementary material state this boundary explicitly.}

\subsection{Adaptive Layer Budget and Selective Global Protection}

A client-level pruning ratio specifies how much capacity a client can retain, but it does not determine where this capacity should be preserved. This is important because different layers do not contribute equally after pruning: some layers contain features that are broadly useful across clients, whereas others can absorb more compression with less impact on the final prediction. FedSAP therefore treats each client's pruning ratio as a total-capacity budget, rather than as a fixed layer-wise pruning rule. For client $i$, the pruning ratio $\rho_i$ gives the target keep ratio $\kappa_i=1-\rho_i$. Instead of applying the same $\kappa_i$ to every layer, FedSAP allocates this budget across layers according to estimated utility.

For each channel, FedSAP defines a general utility from two signals: the activation magnitude, which reflects how frequently the channel responds to local samples, and the gradient magnitude of the corresponding channel weights, which reflects how strongly the channel contributes to local optimization. The two normalized scores are combined as
\begin{equation}
\label{eq:channel_utility}
    u_{i,\ell c}^{(t)}
    =
    \lambda_a \widetilde{r}_{i,\ell c}^{(t)}
    +
    \lambda_g \widetilde{h}_{i,\ell c}^{(t)} .
\end{equation}
In the reported implementation, $\lambda_a=0.5$ and $\lambda_g=0.5$, so utility combines the normalized activation and gradient magnitudes with equal weight. The server averages channel utilities across clients to obtain layer-level importance scores. Layers with higher utility receive larger active-channel budgets, while lower-utility layers receive smaller ones. The resulting keep ratios are normalized to satisfy Eq.~\eqref{eq:budget_constraint}. FedSAP uses a softmax temperature of $1.0$ over layer utilities. Before budget normalization, the per-layer keep ratio is clipped between $\max(0.05,0.15\kappa_i)$ and $\min(1,\kappa_i+0.5)$.

After deciding how many channels each layer can keep, FedSAP determines how the active channels should be shared across clients. This step, called Selective Global Protection, separates retained channels into \emph{Global} and \emph{Private} roles. The key criterion is cross-domain stability. For each pseudo-domain group, the server aggregates the mean activation magnitude of each channel and compares its behavior across groups. We measure the dispersion of channel $c$ in layer $\ell$ as
\begin{equation}
\label{eq:channel_dispersion}
    d_{\ell c}^{(t)}
    =
    \frac{
    \mathrm{Var}_{k}\left(\mu_{k,\ell c}^{(t)}\right)
    }{
    \left(\frac{1}{K}\sum_{k=1}^{K}\mu_{k,\ell c}^{(t)}\right)^2+\epsilon
    },
\end{equation}
where $\mu_{k,\ell c}^{(t)}$ is the average activation magnitude of this channel in pseudo-domain group $k$. Small dispersion indicates similar cross-domain behavior and favors Global sharing; large dispersion indicates domain sensitivity and favors Private preservation. FedSAP smooths these scores with an exponential moving average before role assignment.

For each client and layer, FedSAP first selects the top $\lceil \kappa_{i,\ell}^{(t)} C_\ell\rceil$ channels by utility score and marks the rest as \emph{Dropped}. Among active channels, those with stable cross-domain behavior are assigned to the \emph{Global} pool, while domain-sensitive channels are assigned to the \emph{Private} pool. We also enforce a target global-channel ratio $g_{\mathrm{target}}$ to prevent useful shared features from being overly fragmented into private pools, especially in early rounds with noisy domain estimates.

During local optimization, each client trains its masked model with the standard classification loss and a lightweight representation regularizer:
\begin{equation}
\label{eq:local_objective}
    \mathcal{L}_i
    =
    \mathcal{L}_{\mathrm{CE}}(\mathbf{w}_i;\mathcal{B}_i)
    +
    \frac{\gamma}{|\mathcal{B}_i|}
    \sum_{\mathbf{x}\in\mathcal{B}_i}
    \left\|
    f_{\mathrm{enc}}(\mathbf{x};\mathbf{w}_i)
    \right\|_2^2 .
\end{equation}
Here, $\mathcal{B}_i$ is a local mini-batch, $f_{\mathrm{enc}}$ denotes the encoder output, and $\gamma$ controls the regularization strength. We set $\gamma=0.01$ in all reported experiments.

\subsection{Type-Matched Aggregation and Training Procedure}

After local training, clients upload masked model updates and channel statistics. Since clients may keep different channel subsets and assign active channels to different roles, FedSAP performs type-matched aggregation rather than direct FedAvg~\citep{mcmahan2017communication}. Global channels are aggregated across all clients that keep them as Global, while Private channels are aggregated only within the corresponding pseudo-domain group; Dropped channels remain inactive. The aggregation rule is:
\begin{equation}
\label{eq:type_matched_aggregation}
\mathbf{W}_{\ell c}^{(t)}
=
\begin{cases}
\sum_{i\in \mathcal{C}_{\ell c}^{\mathrm{G}}}
\alpha_i^{\mathrm{G}}\mathbf{w}_{i,\ell c}^{(t)},
&
\text{if channel }(\ell,c)\text{ is Global},\\[0.4em]
\sum_{i\in \mathcal{C}_{k,\ell c}^{\mathrm{P}}}
\alpha_i^{\mathrm{P}}\mathbf{w}_{i,\ell c}^{(t)},
&
\text{if channel }(\ell,c)\text{ is Private},
\end{cases}
\end{equation}
where $\mathcal{C}_{\ell c}^{\mathrm{G}}$ contains clients that keep channel $(\ell,c)$ as Global, and $\mathcal{C}_{k,\ell c}^{\mathrm{P}}$ contains clients in pseudo-domain group $k$ that keep this channel as Private. The weights $\alpha_i^{\mathrm{G}}$ and $\alpha_i^{\mathrm{P}}$ are normalized by data size within each aggregation set.

\rev{Before the next communication round, the server constructs each client-specific model by combining globally shared parameters with the Private parameters assigned to its pseudo-domain group as follows:}
\begin{equation}
\label{eq:client_model_construction}
    \mathbf{W}_{i}^{(t)}
    =
    \mathcal{M}_{i,\mathrm{G}}^{(t)} \odot \mathbf{W}_{\mathrm{G}}^{(t)}
    +
    \mathcal{M}_{i,\mathrm{P}}^{(t)} \odot
    \mathbf{W}_{\pi^{(t)}(i),\mathrm{P}}^{(t)} .
\end{equation}
This construction ensures that all clients receive shared cross-domain knowledge through Global channels, while domain-sensitive knowledge is preserved within pseudo-domain groups through Private channels. Compared with reconstructing all pruned models into a single aggregation path, type-matched aggregation avoids mixing channel updates with incompatible roles.

\begin{algorithm}[t]
\caption{Federated learning with Structured Adaptive Partitioning (FedSAP)}
\label{alg:fedsap}
\begin{algorithmic}[1]
\REQUIRE Initial model $\mathbf{W}^{(0)}$, clients $\{1,\ldots,N\}$, rounds $T$, local epochs $E$, pruning ratios $\{\rho_i\}_{i=1}^{N}$, DGA period $R_d$
\ENSURE Global parameters $\mathbf{W}_{\mathrm{G}}^{(T)}$ and private parameters $\{\mathbf{W}_{k,\mathrm{P}}^{(T)}\}_{k=1}^{K}$
\STATE Set $\kappa_i=1-\rho_i$ and initialize active channels as \emph{Global}.
\FOR{$t=1$ to $T$}
    \STATE Server broadcasts client-specific models and tri-state masks.
    \FOR{each client $i$ in parallel}
        \STATE Train the masked model for $E$ epochs using Eq.~\eqref{eq:local_objective}.
        \STATE Upload model updates, first-layer gradient $\mathbf{g}_i^{(t)}$, utilities $u_{i,\ell c}^{(t)}$ in Eq.~\eqref{eq:channel_utility}, and activation statistics.
    \ENDFOR
    \IF{$t \bmod R_d = 1$}
        \STATE Compute $S_{ij}^{(t)}$ by Eq.~\eqref{eq:dga_similarity} and infer pseudo-domain groups $\{\mathcal{G}_{k}^{(t)}\}_{k=1}^{K}$.
    \ENDIF
    \STATE Aggregate utilities into layer-level importance scores.
    \FOR{each client $i$}
        \STATE Convert $\kappa_i$ into $\{\kappa_{i,\ell}^{(t)}\}_{\ell=1}^{L}$ and normalize them by Eq.~\eqref{eq:budget_constraint}.
        \STATE Select top $\lceil \kappa_{i,\ell}^{(t)} C_\ell\rceil$ channels as active; mark the rest as \emph{Dropped}.
        \STATE Compute dispersion $d_{\ell c}^{(t)}$ by Eq.~\eqref{eq:channel_dispersion} and assign active channels to \emph{Global} or \emph{Private}.
        \STATE Update the tri-state mask in Eq.~\eqref{eq:tri_state_mask} with target global ratio $g_{\mathrm{target}}$.
    \ENDFOR
    \STATE Perform type-matched aggregation by Eq.~\eqref{eq:type_matched_aggregation}.
    \STATE Construct client-specific models by Eq.~\eqref{eq:client_model_construction}.
\ENDFOR
\STATE \textbf{return} $\mathbf{W}_{\mathrm{G}}^{(T)}$ and $\{\mathbf{W}_{k,\mathrm{P}}^{(T)}\}_{k=1}^{K}$
\end{algorithmic}
\end{algorithm}

\begin{table*}[t]
\centering
\small
\setlength{\tabcolsep}{3.2pt}
\renewcommand{\arraystretch}{1.08}
\resizebox{\textwidth}{!}{
\begin{tabular}{lccccccccccc}
\toprule
\multirow{2}{*}{\textbf{FL frameworks}} 
& \multirow{2}{*}{\textbf{System Heter.}} 
& \multicolumn{5}{c}{\textbf{Digits}} 
& \multicolumn{5}{c}{\textbf{Office-Caltech}} \\
\cmidrule(lr){3-7} \cmidrule(lr){8-12}
& 
& \textbf{MNIST} & \textbf{USPS} & \textbf{SVHN} & \textbf{SYN} & \textbf{Global Acc.}
& \textbf{Caltech} & \textbf{Amazon} & \textbf{Webcam} & \textbf{DSLR} & \textbf{Global Acc.} \\
\midrule
FedAvg~\citep{mcmahan2017communication} 
& \xmark 
& 95.89 & 86.84 & 78.39 & 33.63 & 71.81
& 66.07 & 76.84 & 65.52 & 56.67 & 64.54 \\

MOON~\citep{li2021model}                
& \xmark 
& 93.03 & 78.38 & 84.45 & 25.97 & 69.44
& 65.62 & 75.79 & 72.41 & 53.33 & 61.86 \\

FedSR~\citep{caldarola2022fedsr}        
& \xmark 
& 96.77 & 86.15 & 81.48 & 31.64 & 73.89
& 62.95 & 78.95 & 75.86 & 50.00 & 65.47 \\

FPL~\citep{huang2023fpl}
& \xmark 
& 95.54 & 87.69 & 83.74 & 34.73 & 74.17
& 63.84 & 82.63 & 65.52 & 60.00 & 65.45 \\

FedSPU~\citep{niu2025fedspu}            
& \xmark 
& 96.53 & 85.23 & 77.99 & 39.00 & 73.44
& 66.07 & 73.68 & 53.45 & 46.67 & 59.68 \\
\midrule
FedDrop~\citep{bouacida2021adaptivefederateddropout}        
& \cmark 
& 89.48 & 82.51 & 72.98 & 29.35 & 66.85
& 66.07 & 79.47 & 56.90 & 53.33 & 60.58 \\

FedProx~\citep{li2020federated}         
& \cmark 
& 96.68 & 83.96 & 76.69 & 30.95 & 70.74
& 61.61 & 71.05 & 68.97 & 46.67 & 62.08 \\

FedMP~\citep{jiang2022fedmp}         
& \cmark 
& 94.16 & 85.30 & 81.37 & 35.12 & 72.29
& 65.62 & 75.79 & 56.90 & 66.67 & 62.34 \\

NeFL~\citep{kang2025nefl}           
& \cmark 
& 84.98 & 88.49 & 78.41 & 36.02 & 67.64
& 54.91 & 71.05 & 77.59 & 66.67 & 62.26 \\

DapperFL~\citep{jia2024dapperfl}        
& \cmark 
& 96.25 & 86.30 & 82.45 & 37.26 & 74.30
& 64.73 & 81.58 & 74.14 & 66.67 & 67.75 \\

FedSAP (Ours)                           
& \cmark 
& 96.83 & 90.23 & 76.51 & 43.85 & \textbf{76.00}
& 75.45 & 79.47 & 70.69 & 73.33 & \textbf{72.67} \\
\bottomrule
\end{tabular}
}
\caption{\rev{Mean model accuracy (\%) over three random seeds on Digits and Office-Caltech. FedSAP uses $g_{\mathrm{target}}=0.9$, $\tau_{\mathrm{var}}=0.05$, and $\beta_{\mathrm{EMA}}=0.5$. Per-domain entries are best-so-far accuracies and may be attained at different communication rounds; Global Acc. is the best-so-far unweighted mean of domain accuracies evaluated at the same round, so it need not equal the mean of the per-domain columns.}}
\label{tab:main_results}
\end{table*}

\begin{figure*}[!t]
    \centering
    \includegraphics[width=0.8\textwidth]{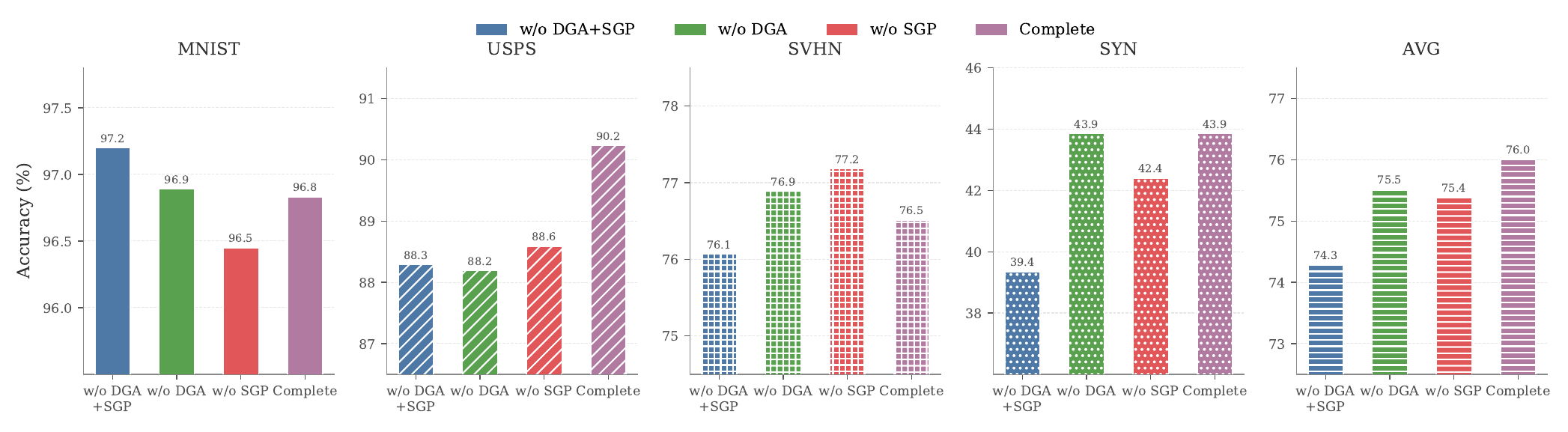}
    \vspace{-0.5em}
    \caption{
    Component ablation of FedSAP on Digits under the selected adaptive configuration.
    Removing DGA, SGP, or both degrades accuracy, showing the complementary benefits of pseudo-domain grouping and stable global-channel protection.
    }
    \label{fig:component_ablation}
    \vspace{-0.8em}
\end{figure*}

\begin{figure*}[!t]
    \centering
    \includegraphics[width=0.7\textwidth]{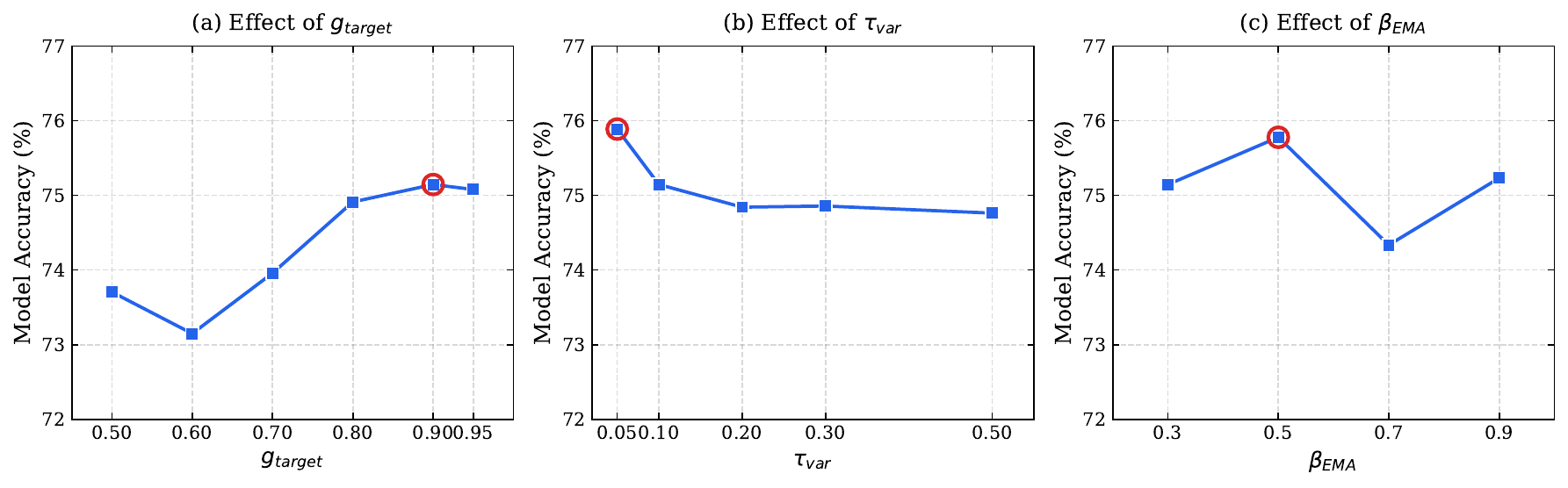}
    \vspace{-0.5em}
    \caption{
    One-factor hyperparameter sensitivity of FedSAP on Digits under the fixed-mask, uniform-budget reference setting.
    The coordinate-wise best values are transferred to the final adaptive configuration.
    }
    \label{fig:hyperparameter_sensitivity}
    \vspace{-0.8em}
\end{figure*}

\section{Experiments}
\label{sec:experiments}

\subsection{Experimental Setup}

\noindent\textbf{Datasets.}
We evaluate FedSAP on two standard multi-domain benchmarks. The first is Digits, which contains four digit-recognition domains: MNIST~\citep{lecun1998gradient}, USPS~\citep{hull1994database}, SVHN~\citep{netzer2011reading}, and synthetic digits. Following the protocol of DapperFL~\citep{jia2024dapperfl}, we distribute 10 clients across the four domains and use $1\%$ of the local data for each client. The second benchmark is Office-Caltech, which contains four object-recognition domains: Caltech~\citep{griffin2007caltech}, Amazon, Webcam, and DSLR~\citep{saenko2010adapting}. We use $20\%$ of each client's local data, following the same protocol. Representative domain examples are shown in Figure~\ref{fig:domain_shift_examples}.
They show that the benchmarks exhibit visible style and acquisition shifts across domains, which aligns with the coupled heterogeneity setting studied in this work.

\noindent\textbf{Models and heterogeneous clients.}
We use ResNet10 for Digits and ResNet18 for Office-Caltech. Following prior heterogeneous FL settings~\citep{diao2021heterofl,jia2024dapperfl}, we divide 10 clients among five capability levels with pruning ratios $\{0\%,20\%,40\%,60\%,80\%\}$.

\noindent\textbf{Baselines.}
We compare FedSAP with three groups of methods. The first group comprises standard FL methods, such as FedAvg~\citep{mcmahan2017communication} and FedProx~\citep{li2020federated}. The second group includes federated domain-generalization methods, MOON~\citep{li2021model}, FedSR~\citep{caldarola2022fedsr}, and FPL~\citep{huang2023fpl}. The third group includes heterogeneous or pruning-based FL methods, FedDrop~\citep{bouacida2021adaptivefederateddropout}, FedMP~\citep{jiang2022fedmp}, NeFL~\citep{kang2025nefl}, and DapperFL~\citep{jia2024dapperfl}. 

\noindent\textbf{Training and evaluation.}
\rev{We train for 100 communication rounds with batch size 64 and SGD at learning rate 0.01. FedSAP updates pseudo-domain groups every five rounds and uses $g_{\mathrm{target}}=0.9$, $\tau_{\mathrm{var}}=0.05$, and $\beta_{\mathrm{EMA}}=0.5$. Every Table~\ref{tab:main_results} entry averages three independent seeds. Following DapperFL~\citep{jia2024dapperfl}, we report per-domain Top-1 accuracy and global accuracy: each domain column is the best-so-far accuracy of that domain, whereas Global Acc. is the best-so-far unweighted mean of the domain accuracies evaluated at the same round. Per-domain bests may occur at different rounds, so their mean need not equal the reported Global Acc.; \mbox{this mean does not imply} per-domain dominance. The supplementary Implementation Details section reports the complete configuration.}

\subsection{Main Results}

\noindent\textbf{Superior global accuracy under system heterogeneity and domain shifts.}
\rev{Table~\ref{tab:main_results} compares methods under heterogeneous device capacities, with client pruning ratios ranging from $0\%$ to $80\%$. Averaged over three random seeds, FedSAP obtains the highest global accuracy on both benchmarks. It reaches $76.00\%$ on Digits, $1.70$ percentage points above DapperFL, and $72.67\%$ on Office-Caltech, a $4.92$-point gain. All gains are absolute points. The central result is that FedSAP improves average accuracy under aggressive heterogeneous compression and domain shift.}

\noindent\textbf{\rev{Cross-domain gains and trade-offs.}}
\rev{The global improvement is not uniform across domains. Relative to DapperFL, FedSAP gains $6.59$ points on SYN, $10.72$ on Caltech, and $6.66$ on DSLR, but loses $5.94$ on SVHN, $2.11$ on Amazon, and $3.45$ on Webcam. Other baselines also exceed FedSAP on some individual domains; for example, FPL is $7.23$ points higher on SVHN, and NeFL is $6.90$ points higher on Webcam. This trade-off is expected because a fixed compression budget forces capacity to be redistributed: protecting channels that improve transfer or specialization for one group can reduce the capacity available to another. FedSAP optimizes the reported cross-domain mean rather than worst-domain accuracy, so the main claim is improved global performance at heterogeneous compression rates---not per-domain dominance. The supplementary domain-wise budget experiment further shows that adaptive allocation helps most on SYN, the most budget-sensitive Digits domain.}

\noindent\textbf{Layer-adaptive and role-aware pruning as a better design choice.}
\rev{The gains in the global mean support a more structured view of pruning than client-level compression alone. Under the same budget, globally stable channels reduce estimation variance by pooling data, Private channels reduce cross-domain bias by restricting aggregation, and Dropped channels satisfy the resource constraint. The supplementary Conditional Analysis formalizes this bias--variance--compression interpretation, derives the role-selection condition, and states its assumptions and failure modes without asserting a universal neural-network guarantee.}

\subsection{Ablation Studies}
\label{sec:ablation}


\noindent\textbf{Complementary benefits of DGA and SGP in channel-role aggregation.}
We evaluate Domain-Guided Assignment (DGA) and Selective Global Protection (SGP) on Digits under the selected adaptive configuration. As shown in Figure~\ref{fig:component_ablation}, the complete FedSAP configuration reaches the best average accuracy of $76.0\%$. Removing DGA or SGP yields $75.5\%$ and $75.4\%$, respectively; removing both modules reduces the average further to $74.3\%$. This suggests that the two modules address complementary aspects of channel-role allocation: DGA determines which clients should share Private channels, whereas SGP identifies which channels should remain globally shared. Their domain-wise effects vary: removing SGP slightly improves SVHN but lowers the source-domain average, so FedSAP favors a balanced multi-domain solution over single-domain optimization. 

\noindent\textbf{Low sensitivity to channel-role hyperparameter choices.}
We study three channel-role hyperparameters on Digits: the target global-channel ratio $g_{\mathrm{target}}$, dispersion threshold $\tau_{\mathrm{var}}$, and EMA coefficient $\beta_{\mathrm{EMA}}$. We vary one at a time while fixing the others. Figure~\ref{fig:hyperparameter_sensitivity} shows competitive accuracy across broad ranges.
The best settings are $g_{\mathrm{target}}=0.9$, $\tau_{\mathrm{var}}=0.05$, and $\beta_{\mathrm{EMA}}=0.5$. They favor a large Global pool, strict stability filtering, and moderate temporal smoothing. We use this coordinate-wise selected combination in the main configuration. The supplementary sensitivity table reports every evaluated value and its global accuracy.

\begin{table}[t]
\centering
\small
\setlength{\tabcolsep}{4pt}
\renewcommand{\arraystretch}{1.05}
\begin{tabular}{lcc}
\toprule
Gradient source & Acc. (\%) & $\Delta$ from Conv1 \\
\midrule
Conv1  & 75.54 & -- \\
Layer1 & \textbf{75.76} & +0.22 \\
Layer2 & 75.68 & +0.14 \\
Layer3 & 75.11 & -0.43 \\
\bottomrule
\end{tabular}
\caption{\rev{Sensitivity to the DGA gradient source under the fixed-mask reference setting on Digits.}}
\label{tab:dga_robustness}
\end{table}

\noindent\textbf{\rev{Sensitivity to the gradient source on appearance-driven domains.}} \rev{We examine how the DGA proxy changes with the layer used for gradient similarity. This diagnostic uses the fixed-mask reference setting, whose complete model obtains $75.54\%$ global accuracy on Digits, and isolates the gradient source rather than comparing against the final dynamic-mask configuration. Table~\ref{tab:dga_robustness} shows comparable results for Conv1, Layer1, and Layer2. Layer1 is highest, while Conv1 offers the smallest payload. Thus, this result supports shallow gradients only for appearance-driven domains.}



\section{Conclusion}
\label{sec:conclusion}
\rev{FedSAP reframes heterogeneous federated pruning as a joint decision over layer capacity and sharing scope. Its Global, Private, and Dropped states respectively pool transferable evidence, preserve pseudo-domain specialization, and meet client budgets. Across three seeds, this allocation improves the global mean on Digits and Office-Caltech under client pruning ratios up to $80\%$, although it does not dominate every domain. These results support role-aware capacity allocation while leaving important boundaries for future study.}

\noindent\textbf{\rev{Limitations}}
\label{sec:limitations}
\rev{First, DGA is a heuristic proxy for latent domains, not a domain-identification guarantee. First-layer gradients are well matched to shifts in texture, color, background, or stroke, but may not separate domains defined by high-level semantics, label mechanisms, or long-range context. Second, transmitting per-client shallow gradients and channel statistics introduces privacy exposure beyond ordinary model updates: an honest-but-curious server may infer domain identity, class proportions, or sample attributes, and gradient inversion is possible in some regimes. FedSAP currently provides no formal privacy guarantee; secure aggregation, clipping with noise, statistic quantization, and attack-based privacy audits are necessary extensions. Third, the tri-state interpretation relies on estimated utility, channel stability, and sufficiently accurate pseudo-domain groups. Incorrect assignments can replace useful global transfer with private estimation noise, or mix incompatible private updates. Finally, our evidence is limited to two vision benchmarks and ResNet-style backbones. The reported global mean can also hide worst-domain regressions, as Table~\ref{tab:main_results} shows. Evaluation on semantic domain shifts, larger architectures, \mbox{partial participation}, imbalanced domains, and worst-domain or fairness-aware objectives is therefore needed. The supplementary Pseudo-Domain Error analysis bounds grouping error under stated assumptions, while Privacy Risk of Uploaded Signals discusses leakage channels, compatible defenses, and attack-based auditing requirements.}
\bibliography{references}

\clearpage
\setcounter{section}{0}
\setcounter{secnumdepth}{2}
\begin{center}
{\LARGE\bfseries Supplementary Material\par}
\end{center}

\appendix

\section{Comparison Frameworks}
\label{app:baselines}

We briefly introduce the comparison frameworks used in our experiments.

\noindent\textbf{FedAvg}~\citep{mcmahan2017communication} is the standard federated learning baseline. Each client performs local training and uploads its model update to the server, which then computes a data-size-weighted average to update the global model.

\noindent\textbf{FedProx}~\citep{li2020federated} extends FedAvg by adding a proximal regularization term to the local objective. The proximal term constrains local models from drifting too far away from the global model, which is useful under statistical heterogeneity.

\noindent\textbf{MOON}~\citep{li2021model} uses model-contrastive learning to reduce representation drift between local and global models. It encourages the current local model to stay close to the global model representation while moving away from previous local representations.

\noindent\textbf{FedSR}~\citep{caldarola2022fedsr} improves federated domain generalization by regularizing latent representations. It penalizes domain-sensitive variations in feature representations and encourages robust features across source domains.

\noindent\textbf{FPL}~\citep{huang2023fpl} introduces prototype-based learning for federated domain generalization. It aligns class-level semantic prototypes across clients and improves generalization under domain shifts.

\noindent\textbf{FedDrop}~\citep{bouacida2021adaptivefederateddropout} reduces local computation and communication cost by training selected subsets of the global model on resource-constrained clients. It is the Adaptive Federated Dropout baseline used in our comparison.

\noindent\textbf{FedMP}~\citep{jiang2022fedmp} uses adaptive model pruning to support heterogeneous edge clients. It adjusts local model footprints according to client resource constraints.

\noindent\textbf{NeFL}~\citep{kang2025nefl} scales nested subnetworks in both depth and width, allowing clients with different resource budgets to participate in FL. It is a representative heterogeneous FL method based on nested model scaling.

\noindent\textbf{DapperFL}~\citep{jia2024dapperfl} is the closest baseline to our setting. It combines Model Fusion Pruning with Domain Adaptive Regularization and reconstructs heterogeneous pruned models before aggregation. It supports clients with different pruning ratios under domain shifts, but does not explicitly allocate channel capacity across layers or separate retained channels by sharing role.

\noindent\textbf{FedSPU}~\citep{niu2025fedspu} is a personalized federated learning method for resource-constrained devices. Instead of training a smaller submodel through dropout-based pruning, FedSPU keeps the full model architecture and randomly freezes a subset of neurons during local training. We include it as an additional resource-constrained FL baseline. Since FedSPU is not designed for federated domain generalization, we evaluate it under the same multi-domain client split and report its global accuracy for comparison.
\section{\rev{FedSAP Settings}}
\label{app:evaluation_config}

We use two FedSAP configurations in the experiments. Both follow the same heterogeneous-device protocol, where clients are assigned pruning ratios from $\{0,0.2,0.4,0.6,0.8\}$. The main configuration uses adaptive layer budgets, dynamic masks, and the coordinate-wise selected channel-role parameters. The diagnostic reference uses uniform layer budgets, fixes masks after warm-up, and anchors the one-factor sweeps at their original values.

\noindent\textbf{Main adaptive configuration.}
The main comparison results use the selected adaptive configuration with dynamic masks. This setting enables adaptive layer-wise budget allocation, where the client-level keep ratio is distributed non-uniformly across layers according to layer utility while preserving the overall client-level compression target. Based on the sensitivity study, we use $g_{\mathrm{target}}=0.9$, $\tau_{\mathrm{var}}=0.05$, and $\beta_{\mathrm{EMA}}=0.5$. The tri-state masks are updated throughout training, which allows channel roles to adapt as local models evolve. This is the configuration used for the main FedSAP results.

\noindent\textbf{Fixed-mask reference configuration.}
For diagnostic analyses, we retain the fixed-mask uniform-budget reference configuration used by the original one-factor sweeps. It uses $g_{\mathrm{target}}=0.9$, $\tau_{\mathrm{var}}=0.1$, and $\beta_{\mathrm{EMA}}=0.3$, and fixes the tri-state masks after warm-up. This reference obtains $75.54\%$ global accuracy on Digits. It isolates individual hyperparameters and should not be confused with the final dynamic-mask, adaptive-budget FedSAP configuration, which uses the coordinate-wise selected values and reaches $76.00\%$.

\begin{table*}[t]
\centering
\small
\setlength{\tabcolsep}{4.5pt}
\renewcommand{\arraystretch}{1.08}
\begin{tabular}{lcc}
\toprule
Configuration item & Main adaptive & Fixed-mask reference \\
\midrule
Layer-budget rule 
& \texttt{adaptive\_layer\_budget}=1 
& \texttt{adaptive\_layer\_budget}=0 \\
Layer-wise budget pattern
& Importance-adaptive 
& Uniform \\
$g_{\mathrm{target}}$
& 0.9
& 0.9 \\
$\tau_{\mathrm{var}}$
& 0.05
& 0.1 \\
$\beta_{\mathrm{EMA}}$
& 0.5
& 0.3 \\
Utility weights $(\lambda_a,\lambda_g)$
& $(0.5,0.5)$
& $(0.5,0.5)$ \\
Budget temperature
& 1.0
& N/A \\
Budget clipping
& $[\max(0.05,0.15\kappa_i),\,\min(1,\kappa_i+0.5)]$
& N/A \\
Regularization weight $\gamma$
& 0.01
& 0.01 \\
Mask update
& Dynamic
& Fixed after warm-up \\
Tri-state partitioning
& enabled
& enabled \\
DGA
& enabled
& enabled \\
SGP
& enabled
& enabled \\
Warm-up rounds
& 10
& 10 \\
Reassignment period
& \rev{5}
& \rev{5} \\
Local epochs
& 10
& 10 \\
Batch size
& 64
& 64 \\
\bottomrule
\end{tabular}
\caption{\rev{FedSAP hyperparameter settings used in the main comparison and diagnostic analyses.}}
\label{tab:fedsap_configurations}
\end{table*}

\section{Implementation Details}
\label{app:implementation}
\noindent\textbf{Compute resources.}
All experiments were conducted on a GPU server with two custom-modified NVIDIA RTX 4090 GPUs, each upgraded to 48GB memory. The server has a 24-core QEMU virtual CPU and 62GB system memory. The software environment uses Ubuntu 22.04 with CUDA 12.1, cuDNN 8, Python 3.12, and PyTorch 2.3.0. Each experimental run uses simulated federated clients on the same server. FedSAP introduces additional computation from DGA updates, channel-statistic collection, mask updates, and type-matched aggregation; the overhead is analyzed in Appendix~\ref{app:overhead}.

\noindent\textbf{Datasets and splits.}
We evaluate all methods on Digits and Office-Caltech. Digits contains four domains: MNIST, USPS, SVHN, and SYN. Following DapperFL~\citep{jia2024dapperfl}, we distribute 10 clients across the four domains and use $1\%$ of each client's domain data as its local training set. Office-Caltech contains four domains: Caltech, Amazon, Webcam, and DSLR. Under the same multi-domain federated protocol, we use $20\%$ of each client's domain data as its local training set. Evaluation is performed on held-out domain data that are not used during federated training.

\noindent\textbf{Network architectures.}
For Digits, we use a ResNet10 variant with four residual stages. For Office-Caltech, we use ResNet18. Both networks include batch normalization and ReLU activations. The classifier dimension is adjusted according to the number of classes in each benchmark.

\noindent\textbf{Federated training protocol.}
\rev{Unless otherwise specified, we use 10 clients with full client participation. Each client trains locally for 10 epochs per communication round with a batch size of 64. We use SGD with momentum $0.9$ and weight decay $5\times 10^{-4}$. The pruning ratios are assigned cyclically from $\{0,0.2,0.4,0.6,0.8\}$ to simulate heterogeneous device capacities. A pruning ratio of $\rho_i$ corresponds to a target keep ratio $\kappa_i=1-\rho_i$. Main-table entries are arithmetic means over three independent random-seed runs; no single-run result is reported as the main comparison.}

\noindent\textbf{Domain-Guided Assignment (DGA).} \rev{DGA computes client similarity using the gradient of the first convolutional layer. Each client uploads this gradient after local training, and the server computes pairwise cosine similarity $S_{ij}^{(t)}$. We threshold $S_{ij}^{(t)}$ with $\tau_{\mathrm{sim}}=0.5$ and take connected components to obtain pseudo-domain groups. Agglomerative clustering on the distance $1-S_{ij}^{(t)}$ is an alternative implementation. We denote the group assignment of client $i$ by $\pi^{(t)}(i)\in\{1,\ldots,K\}$ and update the grouping every $R_d=5$ communication rounds.}

\noindent\textbf{Adaptive layer budget allocation.}
For each client, FedSAP converts the client-level keep ratio $\kappa_i$ into layer-wise keep ratios. The layer budget is determined by global layer utility, which is computed from channel-wise activation-magnitude and gradient-magnitude statistics aggregated across clients. We set $(\lambda_a,\lambda_g)=(0.5,0.5)$ and use a softmax temperature of $1.0$. Before enforcing the channel-weighted budget constraint, each layer keep ratio is clipped to $[\max(0.05,0.15\kappa_i),\min(1,\kappa_i+0.5)]$. The final normalization preserves the client-level keep ratio $\kappa_i$.

\noindent\textbf{Selective Global Protection (SGP)} assigns active channels to Global or Private states according to channel importance and cross-domain activation stability. In the selected adaptive configuration, we use $g_{\mathrm{target}}=0.9$, $\tau_{\mathrm{var}}=0.05$, and $\beta_{\mathrm{EMA}}=0.5$. Here, $g_{\mathrm{target}}$ controls the target proportion of Global channels among active channels, $\tau_{\mathrm{var}}$ is the dispersion threshold for distinguishing stable and domain-sensitive channels, and $\beta_{\mathrm{EMA}}$ is the EMA coefficient used to smooth channel statistics.

\noindent\textbf{Type-Matched Aggregation.}
During aggregation, Global channels are aggregated across all clients, while Private channels are aggregated only within their corresponding pseudo-domain groups. Dropped channels are not aggregated. This prevents domain-invariant and domain-specific channels from being forced into the same aggregation path.
\begin{table*}[t]
\centering
\small
\begin{tabular}{lll}
\toprule
Type & Hyperparameter & Value \\
\midrule
\multirow{4}{*}{FL training}
& Number of clients & 10 \\
& Client participation rate & 100\% \\
& Communication rounds & 100 \\
& Local epochs & 10 \\
\midrule
\multirow{4}{*}{Optimization}
& Batch size & 64 \\
& Optimizer & SGD \\
& Momentum & 0.9 \\
& Weight decay & $5\times 10^{-4}$ \\
\midrule
\multirow{3}{*}{System heterogeneity}
& Pruning ratios & $\{0,0.2,0.4,0.6,0.8\}$ \\
& Target keep ratio & $\kappa_i=1-\rho_i$ \\
& Client assignment & cyclic over pruning ratios \\
\midrule
\multirow{2}{*}{DGA}
& Similarity threshold $\tau_{\mathrm{sim}}$ & 0.5 \\
& Reclustering interval $R_d$ & 5 \\
\midrule
\multirow{4}{*}{Adaptive layer budget}
& Utility weights $(\lambda_a,\lambda_g)$ & $(0.5,0.5)$ \\
& Softmax temperature & 1.0 \\
& Lower clipping bound & $\max(0.05,0.15\kappa_i)$ \\
& Upper clipping bound & $\min(1,\kappa_i+0.5)$ \\
\midrule
Optional regularizer
& Representation weight $\gamma$ & 0.01 \\
\bottomrule
\end{tabular}
\caption{General training hyperparameters used in our experiments. FedSAP-specific configuration differences are shown in Table~\ref{tab:fedsap_configurations}.}
\label{tab:default_hyperparameters}
\end{table*}

\section{Extended Ablation and Hyperparameter Sensitivity}
\label{app:extended_ablation}

\noindent\textbf{Interaction between layer budgeting and aggregation.}
Table~\ref{tab:budget_aggregation_interaction} separates the effect of the layer-budget rule from the effect of tri-state type-matched aggregation. The adaptive budget improves both aggregation variants, while tri-state aggregation provides an additional gain under either budget rule.

\begin{table*}[t]
\centering
\small
\setlength{\tabcolsep}{7pt}
\renewcommand{\arraystretch}{1.12}
\begin{tabular}{lcc}
\toprule
Aggregation path & Uniform layer budget & Adaptive layer budget \\
& (\texttt{adaptive\_layer\_budget}=0) & (\texttt{adaptive\_layer\_budget}=1) \\
\midrule
Single path (w/o Tri-State) & 73.5 & 74.08 \\
Tri-state type-matched (FedSAP) & 75.2 & \textbf{76.00} \\
\bottomrule
\end{tabular}
\caption{Global accuracy (\%) on Digits under the $2\times2$ combination of aggregation path and layer-budget rule. Entries are best-so-far global accuracies over training rounds, averaged over three seeds.}
\label{tab:budget_aggregation_interaction}
\end{table*}

\noindent\textbf{Where adaptive budgeting helps.}
Table~\ref{tab:uniform_adaptive_domains} gives the corresponding domain-level comparison. The improvement is small on the nearly saturated MNIST and USPS domains, moderate on SVHN, and largest on SYN. This pattern is consistent with the interpretation that the hardest domain is most sensitive to where the limited channel budget is placed.

\begin{table*}[t]
\centering
\small
\setlength{\tabcolsep}{8pt}
\renewcommand{\arraystretch}{1.12}
\begin{tabular}{lccc}
\toprule
Domain & Uniform budget + tri-state & Complete FedSAP & Difference pattern \\
\midrule
MNIST & 96.5 & 96.83 & Small (near saturation) \\
USPS  & 89.5 & 90.23 & Small \\
SVHN  & 75.0 & 76.51 & Moderate \\
SYN   & 42.0 & 43.85 & Largest (most budget-sensitive) \\
\bottomrule
\end{tabular}
\caption{Domain-level accuracy (\%) for uniform-budget tri-state aggregation and the complete adaptive-budget FedSAP configuration. Each entry is the best-so-far accuracy of the corresponding domain, which may be attained at a different communication round; the unweighted mean of the domain entries therefore need not equal the synchronized global accuracy in Table~\ref{tab:budget_aggregation_interaction}.}
\label{tab:uniform_adaptive_domains}
\end{table*}

\noindent\textbf{Hyperparameter sensitivity.}
Table~\ref{tab:hyperparameter_sensitivity_digits} reports the full numerical results of the one-factor sensitivity study on Digits. The diagnostic holds a uniform layer budget and fixed masks while varying one channel-role hyperparameter at a time around $(g_{\mathrm{target}},\tau_{\mathrm{var}},\beta_{\mathrm{EMA}})=(0.9,0.1,0.3)$. The coordinate-wise best values are $0.9$, $0.05$, and $0.5$, respectively; we transfer this combination to the final dynamic-mask, adaptive-budget configuration reported in the main table.

\begin{table*}[t]
\centering
\small
\setlength{\tabcolsep}{6pt}
\renewcommand{\arraystretch}{1.08}
\begin{tabular}{lcc}
\toprule
Hyperparameter & Value & Global accuracy (\%) \\
\midrule
$g_{\mathrm{target}}$ & 0.5 & 73.71 \\
$g_{\mathrm{target}}$ & 0.6 & 73.15 \\
$g_{\mathrm{target}}$ & 0.7 & 73.96 \\
$g_{\mathrm{target}}$ & 0.8 & 74.91 \\
$g_{\mathrm{target}}$ & 0.9 & \textbf{75.54} \\
$g_{\mathrm{target}}$ & 0.95 & 75.08 \\
\midrule
$\tau_{\mathrm{var}}$ & 0.05 & \textbf{75.89} \\
$\tau_{\mathrm{var}}$ & 0.1 & 75.54 \\
$\tau_{\mathrm{var}}$ & 0.2 & 74.84 \\
$\tau_{\mathrm{var}}$ & 0.3 & 74.86 \\
$\tau_{\mathrm{var}}$ & 0.5 & 74.76 \\
\midrule
$\beta_{\mathrm{EMA}}$ & 0.3 & 75.54 \\
$\beta_{\mathrm{EMA}}$ & 0.5 & \textbf{75.78} \\
$\beta_{\mathrm{EMA}}$ & 0.7 & 74.34 \\
$\beta_{\mathrm{EMA}}$ & 0.9 & 75.23 \\
\bottomrule
\end{tabular}
\caption{Hyperparameter sensitivity on Digits under the fixed-mask reference setting.}
\label{tab:hyperparameter_sensitivity_digits}
\end{table*}

\section{\rev{Conditional Analysis of Tri-State Allocation}}
\label{app:theory}

{
\subsection{A Bias--Variance--Compression View}

The purpose of this analysis is to explain when the three states are preferable, rather than to claim a distribution-free guarantee for a non-convex neural network. Consider one channel $c$ and $K$ latent domains. Let $q_k$ be the evaluation weight of domain $k$, let $\theta_{k,c}$ denote its population-optimal channel parameter, and approximate the domain risk locally by
\begin{equation}
R_{k,c}(w_c)
=
R_{k,c}(\theta_{k,c})
+
\frac{h_c}{2}\lVert w_c-\theta_{k,c}\rVert_2^2,
\qquad h_c>0 .
\label{eq:quadratic_domain_risk}
\end{equation}
This separable quadratic model isolates two costs that are otherwise entangled in deep networks: bias from sharing a domain-sensitive channel and variance from estimating separate channel parameters with fewer samples.

For a Global channel, the population minimizer is
$\bar{\theta}_c=\sum_k q_k\theta_{k,c}$, and its excess approximation error is
\begin{equation}
B_c^{\mathrm{G}}
=
\frac{h_c}{2}\sum_{k=1}^{K}
q_k\lVert\theta_{k,c}-\bar{\theta}_c\rVert_2^2 .
\label{eq:global_sharing_bias}
\end{equation}
Let $\widehat{\theta}_{\pi(k),c}$ be the parameter estimated for the pseudo-domain group assigned to domain $k$. A Private channel instead has
\begin{equation}
B_c^{\mathrm{P}}
=
\frac{h_c}{2}\sum_{k=1}^{K}
q_k\lVert\theta_{k,c}-\theta_{\pi(k),c}\rVert_2^2 .
\label{eq:private_group_bias}
\end{equation}
Perfect grouping makes $B_c^{\mathrm{P}}=0$ when one parameter is maintained for each latent domain, whereas merging incompatible domains makes this term positive. Denote the estimation variances of pooled and group-restricted estimates by $V_c^{\mathrm{G}}$ and $V_c^{\mathrm{P}}$. Pooling typically gives $V_c^{\mathrm{G}}\leq V_c^{\mathrm{P}}$ because it uses more samples, while Private estimation can have lower sharing bias.

\noindent\textbf{Proposition 1 (conditional role choice).}
Under the separable quadratic approximation in Eq.~\eqref{eq:quadratic_domain_risk}, unbiased parameter estimates, and an additive active-channel budget, the risk-minimizing role for an active channel is Global when
\begin{equation}
B_c^{\mathrm{G}}-B_c^{\mathrm{P}}
\leq
V_c^{\mathrm{P}}-V_c^{\mathrm{G}},
\label{eq:global_private_condition}
\end{equation}
and Private otherwise. A channel is Dropped when its best attainable reduction in
$B_c^{s}+V_c^{s}$, $s\in\{\mathrm{G},\mathrm{P}\}$, is smaller than the reduction offered by other channels competing for the same budget.

\noindent\textbf{Justification.}
For either active role, the expected excess risk decomposes into approximation bias plus estimation variance. Subtracting the Global and Private risks yields Eq.~\eqref{eq:global_private_condition}. With a cardinality budget and separable channel contributions, selecting active channels by their marginal risk reduction gives the Dropped set. Thus, the three states implement a conditional bias--variance--compression decision: Global channels trade a small sharing bias for lower variance, Private channels accept higher estimation variance to avoid cross-domain bias, and Dropped channels sacrifice low marginal utility to satisfy the resource constraint.

FedSAP operationalizes terms that are not observable in practice. Cross-group activation dispersion proxies for $B_c^{\mathrm{G}}-B_c^{\mathrm{P}}$, activation/gradient utility proxies for the cost of dropping a channel, and $g_{\mathrm{target}}$ prevents noisy estimates from assigning too many channels to small Private groups. The proposition explains the mechanism only to the extent that these proxies preserve the relevant ordering. It does not prove that the learned neural-network masks are globally optimal.

\subsection{Pseudo-Domain Error and the Boundary of DGA}

The preceding benefit depends on the quality of $\pi(k)$. If the per-channel risk is $L_c$-Lipschitz in its parameter and the diameter of candidate group parameters is at most $D_c$, assigning an $\varepsilon$ fraction of domain mass to an incompatible Private group adds at most $\varepsilon L_cD_c$ to the weighted channel risk. This bound makes the failure mode explicit: a noisy grouping can erase the bias reduction that justified the Private state. Repeated grouping and exponential smoothing improve temporal stability, but they do not remove systematic proxy error.

First-layer gradients are plausible domain signals when the shift changes local image statistics. Texture, color, background, illumination, and stroke style directly affect early filters, which matches Digits and much of Office-Caltech. A semantically defined domain can violate this premise. For example, two domains may share the same low-level appearance but differ in object co-occurrence, label semantics, scene context, or the conditional distribution $P(Y\mid X)$. Their first-layer gradients can be similar even when later representations or classifier gradients differ. Conversely, a nuisance appearance change can split clients that should share semantic features. The shallow-layer ablation in the main paper establishes robustness only for the evaluated appearance-driven benchmarks; it does not validate DGA for high-level semantic domains. Multi-layer similarity, representation prototypes, or uncertainty-aware soft grouping are appropriate extensions for that setting.

\subsection{What the Analysis Does and Does Not Establish}

The analysis supplies a mechanism-level explanation for improved average accuracy under domain shift: channel-wise sharing can reduce variance where domains agree, restrict negative transfer where they disagree, and allocate the remaining budget to channels with the largest estimated utility. It also predicts the observed trade-off. Because the active-channel budget is fixed, improving the weighted mean can reallocate capacity away from a particular domain, and pseudo-domain error can amplify that loss. Accordingly, FedSAP targets the global mean used by the evaluation protocol; it offers neither a monotonic improvement for every domain nor a worst-domain guarantee.
}

\section{\rev{Privacy Risk of Uploaded Signals}}
\label{app:privacy}

{
FedSAP avoids transmitting raw examples, but this fact alone does not establish privacy. DGA exposes a client-indexed first-layer gradient, and SGP exposes client- or group-level activation and channel-utility statistics. Prior work shows that gradients can support reconstruction attacks~\citep{zhu2019deep,geiping2020inverting}. In FedSAP, even when exact reconstruction is unsuccessful, an honest-but-curious server could use repeated shallow gradients to infer visual domain identity or coarse sample attributes. Channel activation means and gradient magnitudes may also reveal class imbalance, feature prevalence, or changes in a client's local population. Pseudo-domain membership itself is sensitive metadata in applications where a domain corresponds to a hospital, sensor, region, or demographic group.

The current protocol assumes a trusted server for these auxiliary signals and provides no differential-privacy, cryptographic, or empirical attack-resistance guarantee. Its privacy claim must therefore remain limited to avoiding direct raw-data transfer. The risk depends on batch size, local-epoch aggregation, the dimensionality and precision of the transmitted signal, the server's prior knowledge, and whether signals are observable per client or only after aggregation; communication size is not a proxy for privacy.

Several defenses are compatible with FedSAP but change its utility. Secure aggregation within candidate groups can hide individual statistics but requires a preliminary grouping mechanism. Gradient clipping and calibrated noise can provide differential privacy, although noise can corrupt cosine similarity and destabilize DGA. Quantization, random projection, infrequent updates, and transmitting similarity sketches rather than raw gradients can reduce exposure without constituting a formal guarantee. A complete deployment study should report domain/class inference and gradient-inversion attacks, privacy--utility curves, and the effect of each defense on grouping accuracy and worst-domain performance.
}

\section{Computation and Communication Overhead}
\label{app:overhead}

FedSAP introduces additional operations for DGA, channel-statistic collection, mask update, and type-matched aggregation. We analyze their overhead relative to standard FedAvg under the Digits setting.

\noindent\textbf{Computation overhead.}
The dominant cost still comes from local training. For ResNet10 on $32\times32$ inputs, let $F$ denote the FLOPs of one forward pass. With 10 clients, 10 local epochs, about 30 mini-batches per epoch, and batch size 64, the per-round local training cost is approximately
\begin{equation}
    10 \times 10 \times 30 \times 64 \times 3F,
\end{equation}
where the factor 3 approximates forward and backward computation. \rev{DGA requires an additional gradient-based pass only every reassignment period. Since reassignment is performed every five rounds, its amortized overhead remains small compared with the full local training cost.} In our estimate, the total additional computation of FedSAP is about $3\%$--$5\%$ over FedAvg.

\noindent\textbf{Server-side aggregation overhead.}
Type-matched aggregation adds lightweight mask-based indexing and separate averaging for Global and Private channels. This affects aggregation rather than local model training. Since aggregation is much cheaper than repeated local forward-backward passes, its contribution to the total computation is minor.

\noindent\textbf{Communication overhead.}
\rev{FedSAP does not upload raw data or full additional models. Its extra payload consists of first-layer gradients for DGA, channel-level statistics, and tri-state masks. The gradient is transmitted only at DGA rounds, and the remaining statistics scale with the number of channels rather than parameters. For ResNet10, masks and channel statistics contain on the order of $10^3$ values, whereas a model update contains millions of parameters. The resulting \emph{byte overhead} is therefore small relative to standard model-update transmission. This size comparison is not a privacy guarantee; Appendix~\ref{app:privacy} analyzes the information that these signals may expose.}

\begin{table*}[t]
\centering
\small
\setlength{\tabcolsep}{5pt}
\renewcommand{\arraystretch}{1.08}
\begin{tabular}{lcc}
\toprule
Component & FedAvg & FedSAP overhead \\
\midrule
Local training & 100\% & 0\% additional \\
Type-matched aggregation & 100\% & about $1\%$ \\
Mask update & -- & $<0.1\%$ \\
DGA update & -- & about $3\%$--$5\%$ amortized \\
Evaluation & 100\% & 0\% additional \\
Communication & 100\% & \rev{small channel-level payload} \\
\bottomrule
\end{tabular}
\caption{Qualitative comparison of additional overhead relative to FedAvg.}
\label{tab:overhead}
\end{table*}

\section{Additional Results}
\label{app:additional_results}

\begin{figure*}[t]
  \centering
  \includegraphics[width=\linewidth]{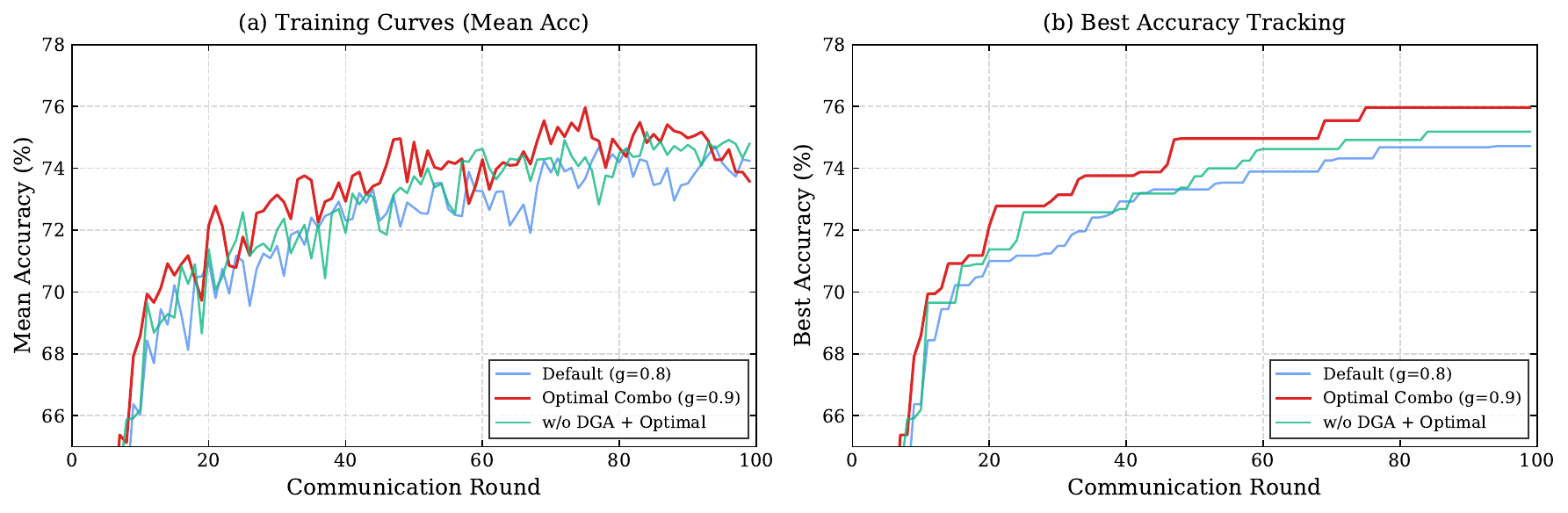}
  \caption{Round-wise and best-so-far global accuracy of three FedSAP configuration variants on Digits.}
  \label{fig:training_curves}
\end{figure*}

\noindent\textbf{Training curves.}
Figure~\ref{fig:training_curves} compares three FedSAP configuration variants used during development: a default $g_{\mathrm{target}}=0.8$ setting, a tuned $g_{\mathrm{target}}=0.9$ setting, and the tuned setting without DGA. Panel (a) reports round-wise mean accuracy, whereas Panel (b) reports the best-so-far accuracy up to each communication round. This is a configuration diagnostic rather than a comparison with external baselines.

\begin{figure*}[t]
  \centering
  \includegraphics[width=\linewidth]{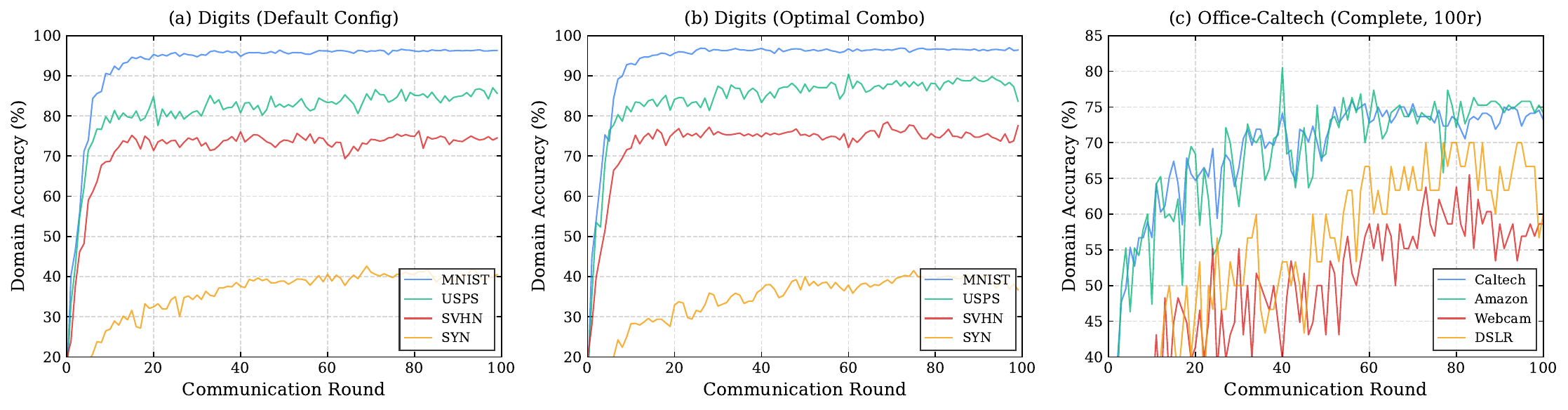}
  \caption{Per-domain accuracy curves on Digits and Office-Caltech.}
  \label{fig:domain_curves}
\end{figure*}

\noindent\textbf{Per-domain accuracy curves.}
\rev{Figure~\ref{fig:domain_curves} shows the domain-wise accuracy curves. The curves expose the same trade-off as the final accuracies: FedSAP improves several visually distinct domains and the cross-domain mean, but does not uniformly improve every domain.}

\begin{figure*}[t]
  \centering
  \includegraphics[width=\linewidth]{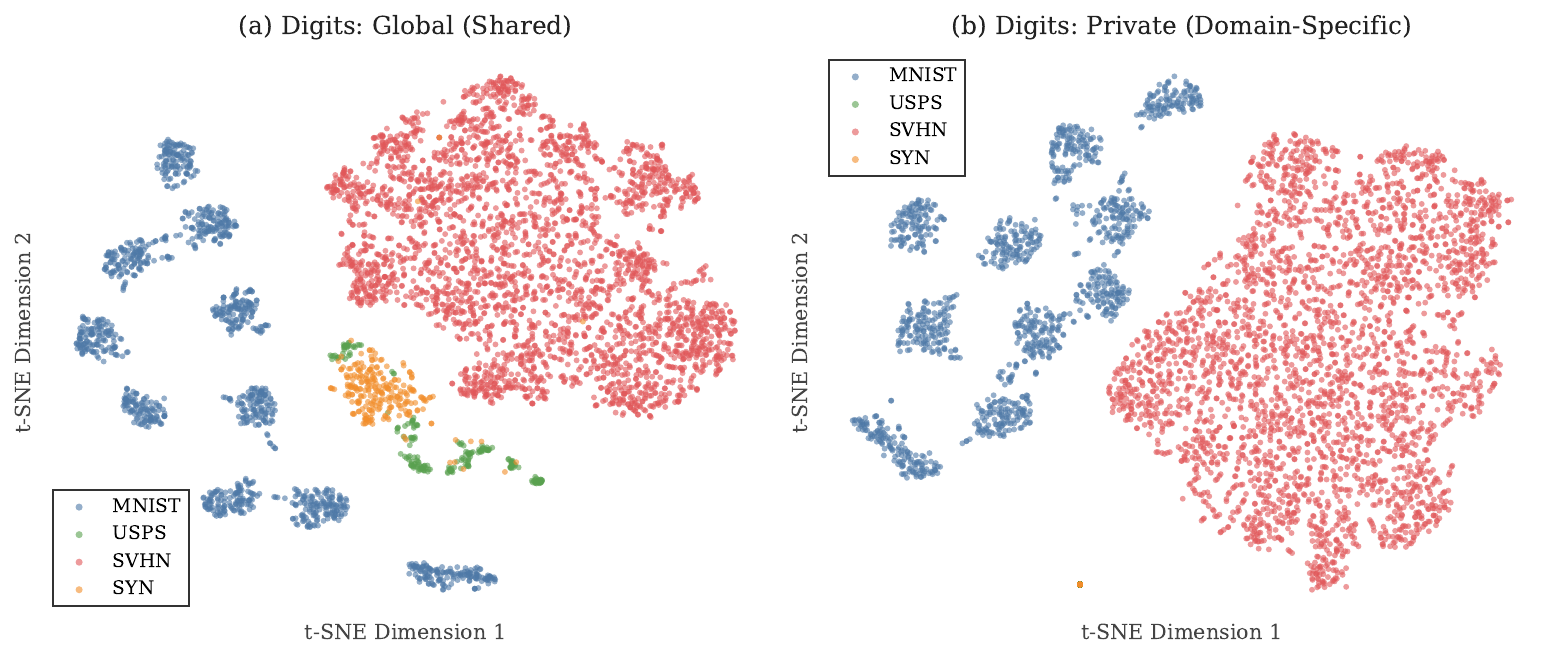}
  \caption{t-SNE visualization on Digits.}
  \label{fig:tsne_digits}
\end{figure*}

\begin{figure*}[t]
  \centering
  \includegraphics[width=\linewidth]{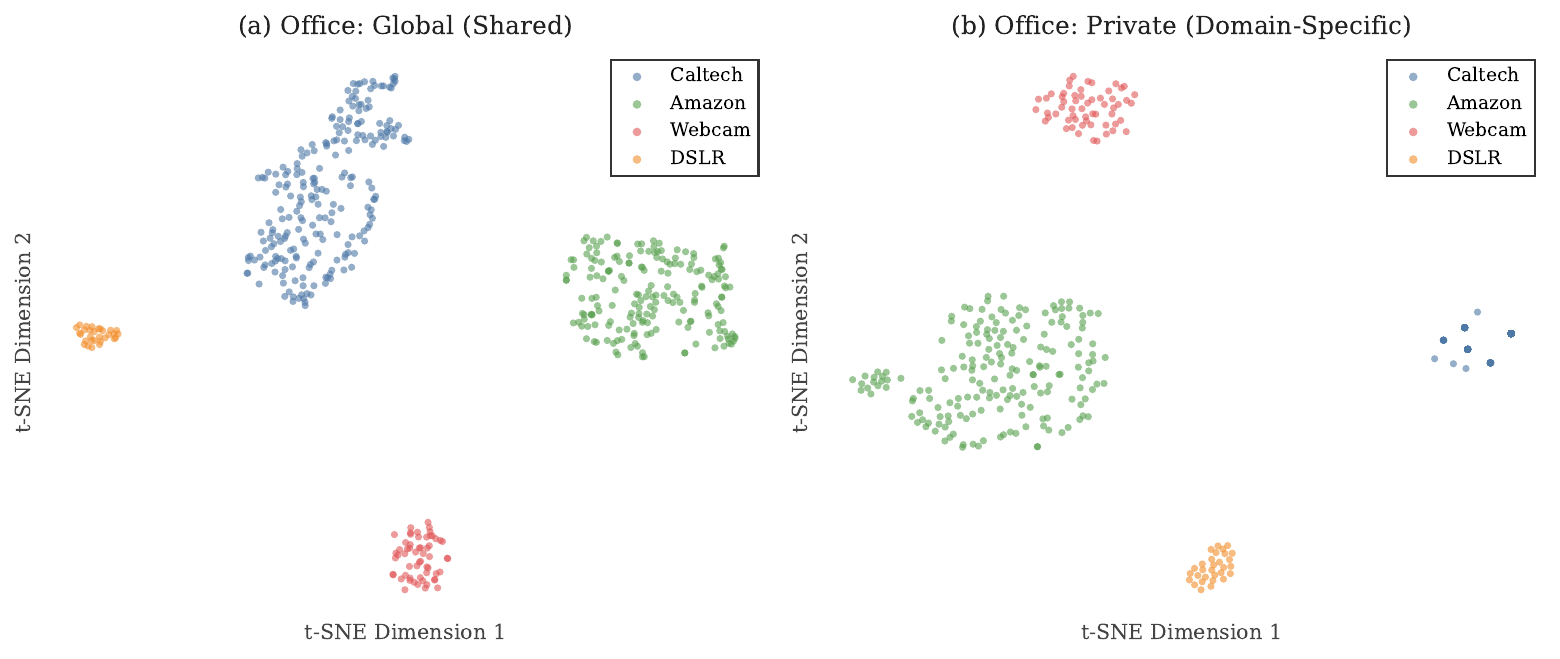}
  \caption{t-SNE visualization on Office-Caltech.}
  \label{fig:tsne_office}
\end{figure*}

\noindent\textbf{Representation visualization.}
Figures~\ref{fig:tsne_digits} and~\ref{fig:tsne_office} show t-SNE visualizations of the learned representations. Global-channel representations show stronger overlap across domains, while Private-channel representations preserve clearer domain-specific structure. This supports the design motivation of role-aware channel allocation.

\end{document}